\documentclass[11pt,hidelinks]{article}

\usepackage[utf8]{inputenc}
\usepackage[T1]{fontenc}
\usepackage{lmodern}
\usepackage{microtype}
\usepackage{orcidlink}

\usepackage[margin=1in]{geometry}

\usepackage{amsmath, amssymb, mathtools}
\usepackage{graphicx}
\usepackage{booktabs}
\usepackage{multirow}
\usepackage{array}

\usepackage{float}     
\usepackage{placeins}  

\usepackage{url}

\usepackage[round]{natbib}
\title{Reflective Translation for Low-Resource Machine Translation via Structured Self-Reflection}
\author{Nicholas Cheng\,\orcidlink{0009-0007-0609-3526}\\Independent Researcher
}

\begin{document}
\maketitle

\begin{abstract}
Low-resource languages such as isiZulu and isiXhosa face persistent challenges in machine translation (MT) due to limited parallel corpora and scarce linguistic resources. Recent work on large language models (LLMs) suggests that self-reflection---the ability of a model to critique and revise its own outputs---can improve reasoning quality and factual consistency. Building on this idea, this paper presents \textit{Reflective Translation}, a prompting framework in which an LLM internally evaluates and corrects its own translations through structured, multi-round prompting to improve semantic fidelity.

The method is evaluated using GPT-3.5 and Claude Haiku 3.5 on English--isiZulu and English--isiXhosa sentence pairs drawn from OPUS-100 and NTREX-African. Translation quality is assessed using BLEU and COMET. Across settings, second-pass translations improve consistently relative to first-pass outputs. This paper further introduces a reflection-augmented dataset consisting of (source, draft, critique, revision) tuples, enabling reproducible analysis of reflective behavior. Overall, the results suggest that reflection-based prompting is a lightweight, model-agnostic approach for improving MT quality in under-resourced languages without fine-tuning or additional labeled data.
\end{abstract}

\section{Introduction}

Machine Translation (MT) enables users to exchange information across languages without human intermediaries. The effectiveness of MT depends on linguistic accuracy, semantic faithfulness, and contextual consistency. Large language models (LLMs) have recently shown strong translation performance without task-specific fine-tuning \citep{brants2007large, moslem-etal-2023-adaptive}. However, a substantial gap remains in low-resource settings \citep{robinson-etal-2023-chatgpt, haddow-etal-2022-survey}, where limited parallel data can lead to hallucinations, omissions, and distortions \citep{wang2020exposure}.

An emerging line of work studies \emph{self-reflection}---prompting models to critique and refine their own outputs---as a mechanism to improve generation quality. Iterative prompting frameworks such as Reflexion \citep{shinn2023reflexion}, Self-Refine \citep{madaan2023selfrefine}, and Chain-of-Verification \citep{creswell2023chainofverification} demonstrate that structured self-evaluation can improve factuality and consistency. Related approaches incorporate reflection signals through training or translation pipelines \citep{li2023reflectiontuning, wang2024reflectionllmmt}.

This paper investigates whether reflection can be applied as an inference-time correction step to improve translation faithfulness without fine-tuning or new labeled data. Translation is treated as constrained reasoning: the target sentence must preserve the meaning of the source. To operationalize this, the proposed framework generates an initial translation, produces a structured self-critique that identifies typical translation errors (mistranslation, omission, semantic distortion), and then produces a revised translation guided by the critique.

\paragraph{Contributions.}
\begin{itemize}
    \item This paper proposes a reflection-guided prompting framework for MT in which models generate and act on structured self-assessments to improve translation faithfulness.
    \item The framework is evaluated on OPUS-100 and NTREX-African for English--isiZulu and English--isiXhosa across two LLMs (GPT-3.5 and Claude Haiku 3.5).
    \item A reflection-augmented dataset of (source, draft, critique, revision) tuples is released to support reproducibility and future analysis.
\end{itemize}

\FloatBarrier

\section{Methods}

\subsection{Reflective Translation}

\textit{Reflective Translation} is a prompting pipeline that guides a model to self-review its own translations and produce improved outputs via structured feedback. For each source sentence, GPT-3.5 \citep{openai2023gpt3.5} and Claude Haiku 3.5 \citep{anthropic2024claude35} generate a first-pass translation. A structured reflection is then produced to identify key errors and provide concise corrective guidance. The model uses this reflection to generate a second-pass translation.

Each reflection consists of:
\begin{enumerate}
    \item \textbf{Error identification:} key mistranslations, omissions, or distortions.
    \item \textbf{High-level fixes:} reusable corrective instructions (e.g., preserve named entities; fix tense/aspect; repair agreement).
    \item \textbf{Critical content:} phrases or semantic constraints that must be preserved.
\end{enumerate}

\subsection{Masking to Reduce Copying}

To reduce leakage from the reflection into the second translation, salient content words are extracted and masked using the Rapid Automatic Keyword Extraction (RAKE) algorithm (NLTK implementation) \citep{rake}. Extracted phrases are replaced with a \texttt{<MASK>} token so the model must apply the critique semantically rather than copy text verbatim.

\begin{figure}[t]
    \centering
    \includegraphics[width=0.85\linewidth]{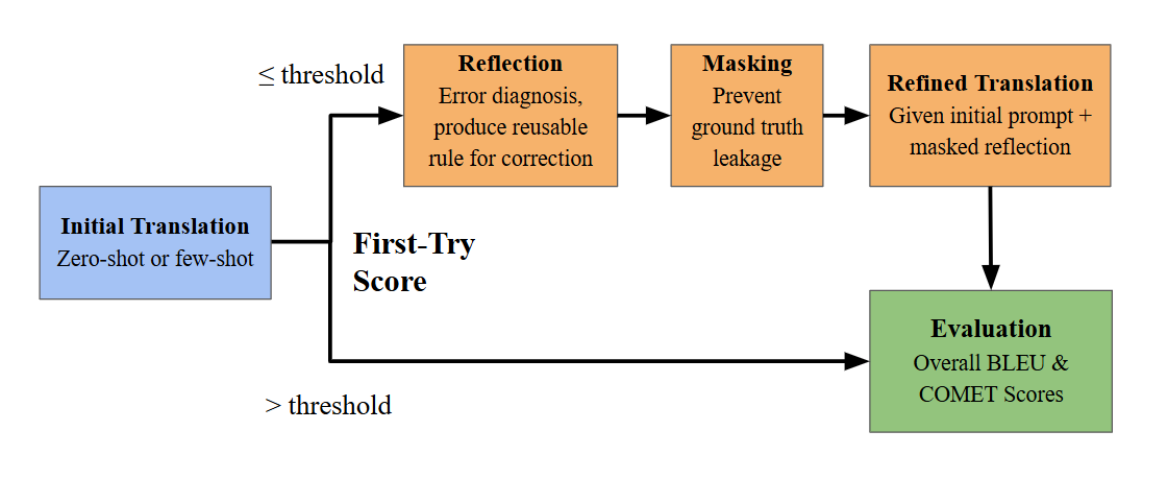}
    \caption{Overview of the reflective translation framework: first-pass translation $\rightarrow$ structured reflection (with masking) $\rightarrow$ second-pass translation.}
    \label{fig:architecture}
\end{figure}

\FloatBarrier

\section{Datasets}

Evaluation uses OPUS-100 \citep{tiedemann2012parallel} and NTREX-African \citep{ntrex2023}. The experiments focus on isiZulu and isiXhosa. OPUS-100 provides broad multilingual parallel data; NTREX-African provides curated evaluation sets for African languages. The experiments use OPUS-100 for English--isiZulu and NTREX-African for English--isiXhosa.

\FloatBarrier

\section{Experimental Setup and Evaluation}

\subsection{Baselines and Prompting Strategies}

Baseline performance uses GPT-3.5 and Claude Haiku 3.5 without fine-tuning. Three prompting strategies are evaluated within the same reflective pipeline:
\begin{itemize}
    \item \textbf{Baseline (zero-shot)} translation.
    \item \textbf{Chain-of-thought-style prompting} \citep{wei2023chainofthoughtpromptingelicitsreasoning} (brief internal reasoning instruction).
    \item \textbf{Few-shot prompting} \citep{brown2020languagemodelsfewshotlearners} with in-context examples.
\end{itemize}

\subsection{Metrics}

Translation quality is assessed using BLEU \citep{papineni2002bleu} and COMET \citep{rei2020comet}. BLEU measures n-gram overlap with a brevity penalty; COMET is a learned metric designed to better track semantic adequacy.

\[
    \text{BLEU} = BP \cdot \exp\left(\sum_{n=1}^{N} w_n \log p_n\right),
\quad
BP =
\begin{cases}
    1 & \text{if } c > r \\
    e^{1-r/c} & \text{if } c \le r
\end{cases}
\]
\[
\text{COMET}(x, y) = f_\theta(x, y)
\]

\FloatBarrier

\section{Results}

\begin{figure}[t]
\centering
\includegraphics[width=0.98\linewidth]{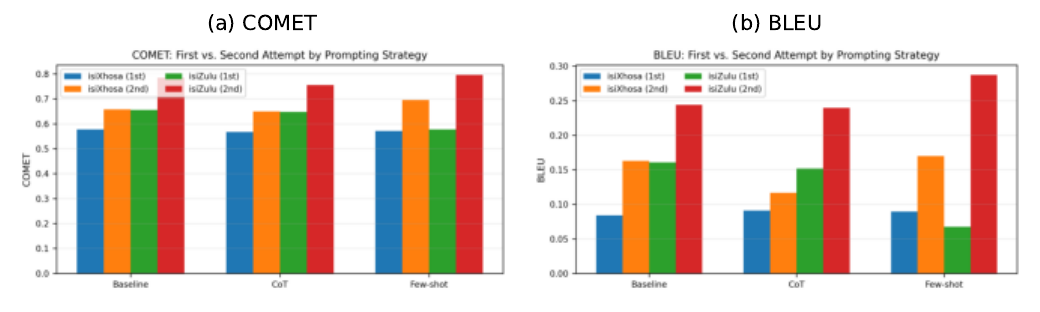}
\caption{First vs.\ second attempt translation quality by prompting strategy. (a) COMET. (b) BLEU.}
\label{fig:prompting}
\end{figure}

Reflective translation improves second-pass outputs across prompting strategies. Figures~\ref{fig:prompting} summarizes the scores for the first vs.\ second attempt by a prompt strategy, showing consistent gains, particularly in COMET.

\begin{figure}[t]
\centering
\includegraphics[width=0.85\linewidth]{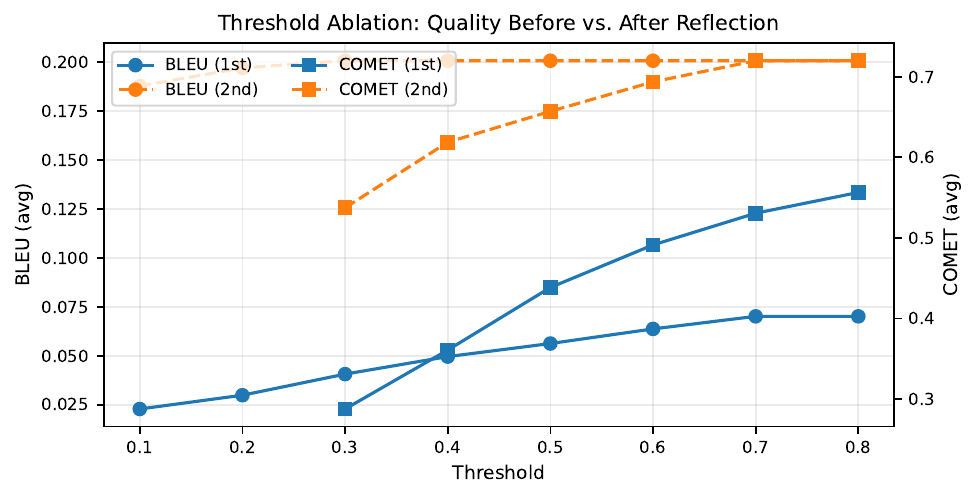}
\caption{Threshold ablation: average BLEU and COMET before vs.\ after reflection across thresholds.}
\label{fig:threshold_ablation}
\end{figure}

Threshold ablation shows that stricter thresholds reduce coverage but can yield larger average gains among refined samples (Figure~\ref{fig:threshold_ablation}).

\FloatBarrier

\section{Data Analysis}

\subsection{First- vs.\ Second-Pass Improvements}

Across both metrics, second-pass outputs improve relative to first-pass translations. COMET improvements are typically larger and more stable than BLEU improvements, suggesting that reflective translation primarily improves semantic adequacy rather than only lexical overlap. This pattern is consistent with the separation observed in Figure~\ref{fig:prompting}, which summarizes first- vs.\ second-attempt scores by prompting strategy, and Figure~\ref{fig:threshold_ablation}, which shows the effect of confidence thresholding.

\subsection{Effect of Confidence Thresholding}

Threshold-based filtering shapes the trade-off between coverage and per-sentence improvement. Higher thresholds reduce the number of sentences eligible for refinement but increase the average improvement among refined samples, consistent with reflective translation functioning as a targeted correction mechanism.

\subsection{Statistical Significance Testing}

To test whether first-to-second pass improvements are statistically significant, paired nonparametric testing is performed between first-pass and second-pass scores at the sentence level. Because BLEU and COMET are not guaranteed to follow a normal distribution and are evaluated on matched sentence pairs, the Wilcoxon signed-rank test is used.

Reflective translation produces statistically significant improvements for both BLEU and COMET. For BLEU, the median paired improvement is +0.0788 over 324 sentence pairs, with the Wilcoxon test strongly rejecting the null hypothesis of zero median difference ($p = 1.45 \times 10^{-44}$). For COMET, the median improvement is +0.1753 over 457 sentence pairs, with $p = 1.10 \times 10^{-65}$.

To quantify practical significance, the rank-biserial correlation associated with the Wilcoxon test is reported. Effect sizes are large for both metrics (BLEU: $r = 0.95$; COMET: $r = 0.96$), indicating that reflective translation yields not only statistically reliable but also practically meaningful gains.

\begin{table}[H]
\centering
\caption{Paired Wilcoxon signed-rank test results comparing first- and second-pass translations.}
\begin{tabular}{lcccc}
\hline
Metric & N & Median Gain & p-value & Effect Size ($r$) \\
\hline
BLEU  & 324 & +0.0788 & $1.45 \times 10^{-44}$ & 0.95 \\
COMET & 457 & +0.1753 & $1.10 \times 10^{-65}$ & 0.96 \\
\hline
\end{tabular}
\label{tab:stat_tests}
\end{table}

\FloatBarrier

\section{Discussion}

The results suggest that structured self-reflection can reliably improve translation quality in low-resource settings without fine-tuning. Improvements are stronger and more consistent in COMET than BLEU, which aligns with reflection correcting semantic errors that may not always increase exact n-gram overlap. Few-shot prompting combined with reflection yields the most stable gains in the prompting comparisons, suggesting that in-context examples can help stabilize the model's self-critique and correction behavior.

\FloatBarrier

\section{Limitations and Future Work}

The evaluation focuses on isiZulu and isiXhosa; broader testing across typologically diverse low-resource languages is required to assess generality. Only two LLMs are evaluated. While BLEU and COMET capture lexical overlap and semantic adequacy, they may miss sociocultural nuance and fine-grained grammatical distinctions; complementary human evaluation would strengthen conclusions. Future work can expand language coverage, model diversity, and explore using (source, draft, critique, revision) tuples for supervised training.

\FloatBarrier

\section{Code and Data Availability}

All code and analysis scripts are available at:
\url{https://github.com/Nickcheng123/reflective-translation-mt}.
The experiments load OPUS-100 and NTREX-African from their original public sources via HuggingFace Datasets; this paper does not redistribute the raw corpora. The repository includes scripts to reproduce tables, figures, and statistical tests.

\FloatBarrier

\appendix
\section{Prompt Templates}

\subsection{Baseline Translation Prompts}

\begin{figure}[H]
    \centering
    \fbox{\parbox{0.98\linewidth}{
    \textbf{Baseline First-Try Prompt:}\\
    \texttt{
Source (\{lang\_name\}): \{source\_text\}\\
You are a professional translator. Translate the given text accurately into English. Preserve the original meaning, tone, and nuance.\\
Output format (exact):\\
Translation:\\
<START\_TRANSLATION>\\
<your English translation here>\\
<END\_TRANSLATION>\\
Do NOT include any explanations.
}}}
    \caption{Baseline translation prompt for the first attempt.}
\end{figure}

\begin{figure}[H]
    \centering
    \fbox{\parbox{0.98\linewidth}{
    \textbf{Baseline Second-Try Prompt (Reflection-Based):}\\
    \texttt{
Source (\{lang\_name\}): \{source\_text\}\\
You are a professional translator. Based on the following review and reflection, provide an improved translation.\\
Reflection: \{reflection\}\\
Output format (exact):\\
Translation:\\
<START\_TRANSLATION>\\
<your improved English translation here>\\
<END\_TRANSLATION>\\
Do NOT include explanations.
}}}
    \caption{Baseline second attempt with reflection.}
\end{figure}

\subsection{Few-Shot Translation Prompts}

\begin{figure}[H]
    \centering
    \fbox{\parbox{0.98\linewidth}{
    \textbf{Few-Shot First-Try Prompt:}\\
    \texttt{
Source (\{lang\_name\}): \{source\_text\}\\
You are a professional translator. Translate the following text into English accurately.\\
Here are examples for guidance:\\
Source (isiZulu): Ngiyabonga kakhulu.\\
Translation: Thank you very much.\\
Source (isiZulu): Unjani namhlanje?\\
Translation: How are you today?\\
Output format:\\
Translation:\\
<START\_TRANSLATION>\\
<your English translation here>\\
<END\_TRANSLATION>
}}}
    \caption{Few-shot prompt with guiding examples.}
\end{figure}

\subsection{Brief Reasoning (Chain-of-Thought-Style) Prompts}

\begin{figure}[H]
    \centering
    \fbox{\parbox{0.98\linewidth}{
    \textbf{Brief Reasoning First-Try Prompt:}\\
    \texttt{
Translate the following \{lang\_name\} text into English.\\
Before giving the final answer, perform brief internal reasoning. Do NOT reveal your reasoning.\\
Source (\{lang\_name\}): \{source\_text\}\\
Output format:\\
Translation:\\
<START\_TRANSLATION>\\
<your English translation here>\\
<END\_TRANSLATION>
}}}
    \caption{First attempt with brief internal reasoning instruction.}
\end{figure}

\FloatBarrier
\clearpage
\bibliography{references}

\end{document}